\documentclass[10pt]{article}
\usepackage[T1]{fontenc}
\usepackage[letterpaper]{geometry}
\usepackage[none]{hyphenat}

\usepackage{times, url, latexsym, indentfirst, authblk}
\usepackage{nicefrac, enumitem}
\usepackage[frozencache=true,cachedir=.]{minted}
\usepackage{lmodern, varwidth, multirow, afterpage}
\usepackage{algpseudocode,algorithm,algorithmicx} 
\usepackage{listings, amsthm, amssymb}
\usepackage[labelsep=period]{caption}
\usepackage[normalem]{ulem}
\usepackage{graphicx, amsmath, breqn}
\usepackage{booktabs}
 \usepackage{collcell, graphicx, array, verbatim}

\usepackage{tikz}
\usetikzlibrary{matrix,positioning,patterns,fit, calc}

\usepackage{pgfplotstable}
\usepackage{pgfplots, colortbl, pgfkeys}

\PassOptionsToPackage{x11names, table, svgnames,dvipsnames}{xcolor}
\pgfplotsset{compat=1.7,}
\pgfplotsset{
    legend image code/.code={
        \draw [#1] (0cm,-0.1cm) rectangle (0.6cm,0.1cm);
    },
}

\algrenewcommand\algorithmicrequire{\textbf{Precondition:}} 
\algrenewcommand\algorithmicensure{\textbf{Postcondition:}}

\newminted{python}{
		   numbersep=8pt,
		   bgcolor=bg,
		   framesep=3mm} 
  \pgfkeys{/heat/.is family, /heat,
      Max colour/.initial = black,
      Min colour/.initial = white,
      max colour/.initial = black,
      min colour/.initial = white,
      text colour/.initial = black,
      Min color/.style = {Min colour=#1},
      Max color/.style = {Max colour=#1},
      min color/.style = {min colour=#1},
      max color/.style = {max colour=#1},
      text color/.style = {text colour=#1},
      min/.initial = -1,
      max/.initial = 1,
      slider/.code={%
         \tikz{\shade[left color=\HVal{min colour},%
                      right color=\HVal{max colour}]%
            (current page.south west) rectangle ++(#1,12pt);
         }%
      }%
  }
  
  \newcommand\HVal[1]{\pgfkeysvalueof{/heat/#1}}

  \newcolumntype{H}{>{\collectcell\Heat}r<{\endcollectcell}}
  \newcommand\Heat[1]{
    \if\relax\detokenize{#1}\relax
    \else%
      \pgfmathparse{int(100*(#1-\HVal{min})/(\HVal{max}-\HVal{min}))}
      \ifnum\pgfmathresult>100
        \edef\HeatCell{\noexpand\cellcolor{\HVal{Max colour}}}%
      \else\ifnum\pgfmathresult<0
          \edef\HeatCell{\noexpand\cellcolor{\HVal{Min colour}}}%
        \else
          \edef\HeatCell{\noexpand\cellcolor{\HVal{max colour}!\pgfmathresult!\HVal{min colour}}}%
        \fi%
      \fi%
      \HeatCell\textcolor{\HVal{text colour}}{$#1$}%
    \fi%
  }

\pgfplotstableread[col sep=comma]{combined-kappas.csv}{\loadedtable}
\pgfplotstablegetcolsof{\loadedtable}
\pgfmathtruncatemacro{\NoOfCols}{\pgfplotsretval-1}
 \pgfplotsset{
        discard if not symbolic/.style 2 args={
            filter discard warning=false,
            x filter/.append code={
                \edef\tempa{\thisrow{#1}}
                \edef\tempb{#2}
                \ifx\tempa\tempb
                \else
                    \def\pgfmathresult{NaN}
                \fi
            },
        },
        discard if not/.style 2 args={
            filter discard warning=false,
            x filter/.append code={
            \ifdim\thisrow{#1} pt=#2pt
                \else
                    \def\pgfmathresult{NaN}
                \fi
            },
        },
    }

\title{Diagnosis of Acute Poisoning Using Explainable Artificial Intelligence}
\date{$^1$ Weill Cornell Medical Centre, NY, NY\\
     $^2$ Brigham and Women's Hospital, Boston, Massachusetts\\
     $^3$ Boston Children's Hospital, Boston, Massachusetts\\
     $^*$ Corresponding author: mic9189@med.cornell.edu}
\author{Michael Chary, MD PhD$^{1,3*}$, Ed W Boyer, MD PhD$^2$,  Michele Burns, MD, MS$^3$}
\providecommand{\keywords}[1]{\textbf{\textit{Keywords\textemdash }} #1}

\begin{document}
\maketitle
\begin{abstract}
 Medical toxicology is the clinical specialty that treats the toxic effects of substances, for example, an overdose, a medication error, or a scorpion sting. The volume of toxicological knowledge and research has, as with other medical specialties, outstripped the ability of the individual clinician to entirely master and stay current with it. The application of machine learning/artificial intelligence (ML/AI) techniques to medical toxicology is challenging because initial treatment decisions are often based on a few pieces of textual data and rely heavily on prior knowledge, experience, and expertise. ML/AI techniques, moreover, often do not represent knowledge in a way that is transparent for the physician, raising barriers to usability. Rule-based systems are more transparent approaches, but often generalize poorly and require expert curation to implement and maintain. Here, we construct a probabilistic logic network to represent a portion of the knowledge base of a medical toxicologist. Our approach transparently mimics the knowledge representation and clinical decision-making of practicing clinicians and requires minimal maintenance.  The software, dubbed \emph{Tak}, performs comparably to humans on straightforward cases and intermediate difficulty cases, but is outperformed by humans on challenging clinical cases. \emph{Tak} outperforms a decision tree classifier at all levels of difficulty. Probabilistic logic provides one form of explainable artificial intelligence that may be acceptable for use in healthcare.

\end{abstract}

\section{Introduction \label{sec:intro}}
    The goal of this project is to represent physician decision making in acute conditions in medical toxicology using explainable and transparent computational techniques. The scope of biomedical knowledge is too vast and rate of increase of that knowledge too rapid for an individual physician to bring all relevant knowledge to bear on the diagnosis and treatment of an illness. The traditional response is the formation of teams of specialists, but this can promote fragmentation of care and increases the cost of delivering healthcare as well the chances for miscommunication. Nor does specialization remove elements of human fatigue or latent bias. Machine learning and artificial intelligence (ML/AI) approaches are trained on data sets larger than any physician could encounter in training. ML/AI algorithms can outperform physicians on \emph{specific} tasks, such as predicting the likelihood of response to a chemotherapeutic regimen \cite{rajkomar2019machine} or diagnosing pneumonia from a chest X-ray \cite{rajpurkar2017chexnet}, but perform dismally in poorly-defined tasks such as constructing a differential diagnosis, a list of diagnoses ranked by the likelihood of explaining a patient's current condition. 
    
    A barrier to integrating ML/AI into healthcare is the difference between how ML/AI and physicians evaluate clinical data. Many current ML/AI approaches look for quantitative patterns across large data sets. They ignore prior knowledge (\emph{i.e.} information an experience physicians acquire in medical school and residency). Pretrained models and transfer learning incorporate statistical relationships from prior knowledge, but do not explicitly represent these relationships in terms readily interpretable by physicians. Reliability plots and counterfactual reasoning provide a way to understand the internal reasoning of algorithms that classify pictures of biopsies \cite{thiagarajan2020calibrating}, but it is not clear how to apply these methods to \emph{textual} data.  Probabilistic logic provides a way to combine statistical learning with symbolic reasoning, a way to combine machine capacity with human intuition.   
    
\subsection{The Diagnosis and Initial Treatment of a Poisoned Patient\label{sec:toxidromes}}
The diagnosis and treatment of a poisoned patient begins with a rapid determination of whether the patient requires immediate intervention to prevent death. This determination is usually made at the patient's bedside by a physical examination and, if the patient's mental status allows, a brief discussion with the patient. In these critical situations, laboratory tests (\textit{e.g.} serum or urine drug concentrations) are rarely available readily enough to inform this determination. Opioids can slow breathing within minutes of ingestion. The metabolites are not detected in urine for hours, but the effect of the drug needs to be immediately reversed to prevent death from lack of oxygen. Serum concentrations of opioids and sedatives often have to be sent to specialized labs and the results are not available for days to weeks. 

The medical toxicologist relies on pattern of findings on bedside evaluation, termed \emph{toxidromes}, that suggest a life-threatening ingestion from a class of drugs, \emph{e.g.} a sedative or a stimulant. Table \ref{tab:toxidromes} displays the $6$ canonical toxidromes, the features physicians extracts, and the feature values for each toxidrome. The bedside evaluation a medical toxicologist performs resembles, in the language of ML/AI, feature extraction and then multinomial classification with a complex loss function. From this lens, a toxidrome is a set of ranges of values of features that define decision regions for class membership. The word \textit{toxidrome} refers to a decision region. Toxidromes are intended to accurately identify severe poisonings that will respond to treatment, but may misclassify milder poisonings. This misclassification is acceptable clinically because mild poisonings, in general, do not require any specific immediate treatment.

\begin{table}
\setlength\tabcolsep{2pt}
\centering
\begin{tabular}{rcccccccc}
 & HR & BP & Pup & Sec & Temp  & RR & MS \\
Anticholinergic & $\Uparrow$ &  & {\Large$\bullet$} & $\Downarrow$ & $\Uparrow$ &   & D \\
Cholinergic & $\Downarrow$&  & {\tiny$\bullet$} & $\Uparrow$ &  & $\Downarrow$  & S \\
Opioid &  &  & {\tiny$\bullet$}&  &  &  $\Downarrow$ & S \\
Sedative-Hypnotic &  &  &  &  &  &  &  S \\
Serotonin Toxicity & $\Uparrow$ & $\Uparrow$  &  &  & $\Uparrow$  & &  A \\
Sympathomimetic & $\Uparrow$ & $\Uparrow$ & {\Large$\bullet$} &  & $\Uparrow$ & $\Uparrow$ & A
\end{tabular}
\caption{\textbf{Six canonical toxidromes.} HR, heart rate; BP, blood pressure; Pup, pupil diameter, size of bullet represents increased or decreased pupil diameters; Sec, secretions; Temp, temperature; RR, respiratory rate; MS, mental status; D, delirious, S, sedated; A, agitated Empty cell indicates expectation of no abnormality for that sign.}
\label{tab:toxidromes}
\end{table}

The names of the toxidromes reflect the biochemical pathways excessively activated or blocked by classes of drugs. The anticholinergic toxidrome results from blockade of the acetylcholine receptor family; cholinergic toxidrome from activation of the acetylcholine receptor family; opioid toxidrome from activation of the $\mu$ opioid receptors; sedative-hyponotic toxidrome from activation of the GABA ($\gamma$-amino butyric acid) receptors or blockade of glutamate receptors, and the sympathomimetic toxidrome from activation at adrenaline or noradrenaline receptors \cite{holstege2012toxidromes}. Serotonin toxicity is thought to result from excess activation at the 5-HT$_{2A}$ receptor\cite{boyer2005serotonin}. The clinical findings from excess serotonin activation are canonically termed serotonin syndrome or serotonin toxicity, but the term is used equivalently to a toxidrome. 

\subsection{Review of Relevant Studies \label{sec:prior-work}}
Machine learning techniques (\textit{e.g.} naive Bayesian classifiers, neural networks, decision trees) have been applied to  diagnosis in many fields of medicine\cite{kononenko2001machine}. Here we review the application of probabilistic logic networks that analyze \emph{text} to perform medical diagnosis. We exclude algorithms that analyze images, as might be done in radiology or pathology. Images may be helpful in identifying whether flora or fauna are poisonous. Support vector machines have been productively applied to the automatic identification of poisonous mushrooms \cite{wibowo2018classification} and plants \cite{prasvita2013medleaf, sharma2015review} from images. But, image data are often not available to the toxicologist.  

A combination of Bayesian networks and an ontology has been used to diagnose osteoporosis, achieving a $72\%$ accuracy \cite{agarwal2016ontology}. A Markov logic network has been implemented for diagnosis of medical conditions from Chinese-language medical records\cite{jiang2017learning}, but there was no assessment of its performance. The construction of a fuzzy Bayesian network for medical diagnosis has also been proposed but its performance not assessed \cite{zarikas2018medical}.

Software that combined rules and probabilities was developed for medicine as early as the 1970's, for example \textsc{MYCIN}\cite{shortliffe1974mycin}. Research on neural networks eclipsed work on rule-based systems because neural networks could operate with inexact matching, performed more accurately, and scaled more easily and rapidly. The use of neural networks in clinical practice is limited, however, owing, in part, to the difficulty of a clinician interacting with something that ``doesn't speak my language''\cite{naylor2018prospects}.  

To the authors' knowledge there have been no prior publications on the application of probabilistic logic networks to the diagnosis of the poisoned patient. The Chemical Hazards and Emergency Medical Management branch (CHEMM) of the US Department of Health and Human Services has developed the CHEMM Intelligent Syndromes Tool. This tool is available via a web-based interface, but no application programming interface or other scalable endpoint is provided. There are no publications describing its implementation, but it appears to be based on FALCON\cite{frysinger2007falcon}, a deterministic decision tree to co-ordinate responses against attacks with chemical weapons.

\subsection{Probabilistic Logic Networks \label{sec:plns}}
Probabilistic logic networks (PLNs) aim to represent knowledge about the world and allow inference under uncertainty using a combination of predicate logic, symbolic reasoning, and statistical inference\cite{goertzel2008probabilistic}. A PLN consists of a set of pairs of a probability and a logical statement.     

\begin{equation}
    0.4\quad\textrm{somnolent}\left(x\right)\Rightarrow\textrm{sedative_hypnotic}\left(x
    \right)\label{eq:pln-example}
\end{equation}

Equation \eqref{eq:pln-example} represents the concept that in $40\%$ of possible worlds if a patient is somnolent (excessively sleepy and lethargic) then the patient may be poisoned by a medication from the sedative/hypnotic class. 
    
The fraction associated with each statement represents the fraction of words in which the logical statement is true. The statement is assumed to be false in all other worlds. The Supplementary Material provides more detail on the underlying mathematics and our software implementation.  

Rule-based systems need experts to create and curate the rules as well as to adapt the rules to include new knowledge or apply the system to unfamiliar types of data. This need for curation may limit the speed of development. It also provides an opportunity for physicians to contribute to software development, which may increase use in clinical practice. For a more complete introduction to probabilistic logic we refer the reader to \cite{kimmig2011implementation} and for software implementation to \cite{de2003probabilistic}.
   
A significant advantage of rule-based approaches over other machine learning approaches for many areas of medicine is that rule-based approaches do not require large training data sets. Many subspecialties within medicine diagnose and treat rare diseases for which large data sets to explore all methods of diagnosis and treatment are unlikely to exist. The rules, in addition, can be a distillation of the received knowledge of a field, or a combination of this distillation and relationships inferred from large data sets. 

Decision tree (DT) learning provides a competitive alternative to probabilisitic logic networks. DT classifiers have been used to predict the risk of breast cancer\cite{lavanya2012ensemble}, heart disease\cite{soni2011predictive}, diagnose diabetes\cite{al2011decision}, and classify electroencephalogram outputs in patient with epilepsy\cite{polat2007classification}. DTs are robust against collinearity. This is important in toxicology where poisonings share overlapping features. For example an elevated heart rate can be seen in the anithcolinergic and sympathomimetic toxidromes as well as in the serotonin syndrome. The sympathomimetic toxidrome and serotonin syndrome also both have elevated blood pressure.

DTs and PLNs also require less training data than neural networks. This is advantageous for medical applications, where curated data are often tiny. All of the studies discussed above were developed on $150$ or fewer patient presentations.

A limitation of decision trees is the tendency to overfit, which would correspond in clinical practice to bending diagnostic criteria so that each patient has at least one diagnosis. PLNs avoid this overfitting because they are not trying to minimize the number of undiagnosed patients. 

\section{Methods \label{sec:methods}}

\subsection{Knowledge Representation}
   We created $34$ probabilistic logic rules based on the consensus of three medical toxicologists to describe the medical knowledge base used to diagnose acute poisoning. We named these rules and the underlying implementation in ProbLog, \emph{Tak}. We restricted ourselves to developing rules that described features that could be observed during one evaluation at a patient's bedside without laboratory testing. We followed these restrictions to assess our algorithm's performance in the most time-sensitive aspect of medical toxicology. 
   
   The rules were constructed as follows. We treated each finding elicited by the toxicologist as a predicate. A predicate is a function that returns only \textsc{True} or \textsc{False}, dpeending on its input. For example, the predicate \mintinline{prolog}{salivation(X,increased)} is true if patient demonstrates increased salivation. We considered all predicates representing clinical findings to require two inputs, the patient and the value of the feature, usually \{present|absent\} or \{increased|normal|decreased\}. These values reflect a discretization of underlying continuous variables that reflect a common pattern of communication with information compression between physicians. 
   
   \begin{listing}
   \begin{minted}{prolog}
    0.10::salivation(X,decreased);
    0.10::salivation(X,increased);
    0.80::salivation(X,usual).
   \end{minted}
   \caption{\textbf{Example rule in probabilistic logic that represent knowledge from medical toxicology.} Fraction preceding each function denotes number of worlds in which that function is true. Numbers sum to one across values.  \label{lst:prior-probability}}
   \vspace{-1.5em}
   \end{listing}
   
   In Listing \ref{lst:prior-probability}, the number before the two colons represents the probability with which the probability is true. The semicolon represents logical exclusive disjunction, \emph{i.e.} \mintinline{prolog}{A;B} means ``A or B but not both''. A period terminates each logical statement. Listing \ref{lst:eg-posterior-probability} demonstrates assigning the likelihood of one toxidrome over another given that the patient is manifesting a symptom. A colon followed by a dash represents unidirectional implication, \emph{i.e.} \mintinline{prolog}{A :- B} means that B is true if A is true. The function \mintinline{prolog}{mentalStatus(X,agitated)} is true if patient X is agitated. The function \mintinline{prolog}{hasToxidrome(X,Y)} is true if patient X manifests toxidrome Y. 
   
   \begin{listing}
   \begin{minted}{prolog}
    4*P::hasToxidrome(X,sympathomimetic);
    P::hasToxidrome(X,serotonergic) :-
    mentalStatus(X,agitated), P is 0.2.
   \end{minted}
   \vspace{-1.5em}
   \caption{\textbf{Example rule in probabilistic logic linking a symptoms to toxidromes.} Expression preceding each goal in disjunction (sequence of statements separated by semicolons) is evaluated when conditions of goal are satisfied. \label{lst:eg-posterior-probability}}
   \end{listing}

   \begin{listing}
   \begin{minted}{prolog}
    hasToxidrome(X,cholinergic) :- 
    salivation(X, increased),
    urination(X, increased),
    pupilDiameter(X,small).
   \end{minted}
   \vspace{-1.5em}
   \caption{\textbf{Example Expression of Diagnosis of Toxidrome as Prolog Goal.} \label{lst:eg-cholinergic-goal}}
   \end{listing}
 
 The relative probabilities across rules were chosen to reflect the perceived relative prevalence of each clinical finding. We used the most recent annual report from American Association of Poison Control Centers on the relative prevalence of each poisoning in the US to estimate the prior probability of each toxidrome. The prevalence of many physical findings, for example hypersalivation in the general population, are not known. Nor is it known that a patient is exactly four times more likely to suffer from a sympathomimetic toxidrome as opposed to serotonin toxicity if the patient becomes agitated after an unknown ingestion. The magnitudes were chosen, in conjunction with the consensus of experts, to reflect implicit components of clinical reasoning. 
 
\subsection{Data Set}
We generated $300$ simulated toxidrome presentations as follows. We chose $300$ as the minimum number needed to detect whether the inter-rater reliability between human consensus and \emph{Tak} was greater than $0.2$. We took a difference in inter-rater reliability of $0.2$ to indicate clinically significant disagreement. The inter-rater reliability among toxicologists in using toxidromes to diagnose poisonings has not been systematically studied and the need for such a study is acknowledged\cite{larsen2018reviewing}. 

For each presentation we chose two toxidromes according to a uniform random distribution without replacement and a level of difficulty, $k$. One toxidrome was the \emph{intended} toxidrome. The other became the distractor toxidrome. We created the presentation by choosing $5-k$ signs from the intended toxidrome and $k$ signs from the distractor toxidrome. The parameter $k$ models the variability of clinical presentation. A difficulty of $0$ simulates an unequivocal presentation, where all findings are canonically associated with the intended toxidrome. A difficulty of $2$ simulates a mixed picture, as might result from the ingestion of many substances with conflicting effects.  To the author's knowledge, this is the first data set created to evaluate the inter-rater reliability of diagnosis in medical toxicology. 

\subsection{Evaluation}
We compared the performance of \emph{Tak} with expert consensus, a decision tree, and recovery of the intended toxidrome generated by Algorithm \ref{alg:make-patients}. We presented the same $300$ cases described above to \emph{Tak} as well as two human medical toxicologists. \emph{Tak} assigned the most likely rating to each toxidrome based on the posterior probabilities calculated by the probabilistic logic engine. The human raters labeled each presentation with the toxidrome that most accurately captured the presentation. We excluded presentations if either rater felt that no single toxidrome captured the presentation or that the presentation was not clinically plausible. 

We use the term \emph{inferred toxidrome} to denote the toxidrome that \emph{Tak}, the decision tree, or the human raters inferred from the case presentation. We quantified inter-rater reliability using a multinomial extension of Cohen's $\kappa$. Cohen's $\kappa$ ranges between $0$ and $1$ where $0$ indicates the level of agreement expected by chance and $1$ indicates perfect agreement. Cohen's $\kappa$ is more clinically relevant than accuracy because Cohen's $\kappa$ also considers the discordance in errors between raters. We felt it important that \emph{Tak} perform similarly to humans in both accurate and inaccurate diagnoses. 

We used the inter-rater reliability between \emph{Tak}'s inferred toxidrome and the intended toxidrome as a measure of best performance and between \emph{Tak}'s inferred toxidrome and the consensus of the human raters as a measure of actual performance. We took the inter-rater reliability between the consensus of the human raters and the intended toxidrome as a benchmark for actual performance.  

To provide a machine learning benchmark we trained a decision tree classifier on the same cases. We trained one decision tree classifier for each level of difficulty. Training one DT for each level of difficulty is likely to lead to overfitting, but also will overestimate the decision tree's performance, providing a more stringent benchmark against which to evaluate \emph{Tak}. The trees were not averaged nor was any random forest method used. We used the \emph{sklearn} implementation, DecisionTreeClassifier, with the maximum depth set to $3$. 

\section{Results \label{sec:results}}

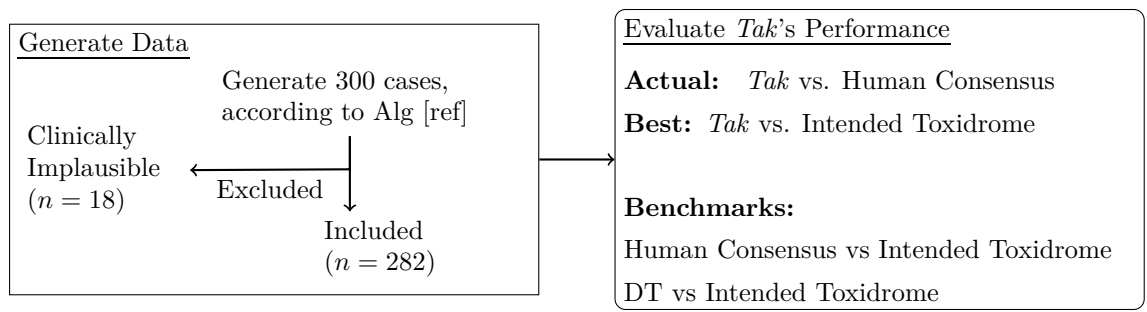
\begin{figure}
    \centering
    \begin{tikzpicture}
        \node[draw,rectangle] (generate-data)
        {
            \begin{minipage}{0.45\textwidth}
            \uline{Generate Data}\\
            \begin{tikzpicture}
                \node[text width=0.5\textwidth] (generate-cases) {Generate $300$ cases,\\ according to Alg [ref]};
                \node[below =  of generate-cases, text width=0.1\textwidth] (included){Included\\$\left(n=282\right)$};
                \node[below left= -0.26cm and 0.30cm of generate-cases, text width=0.3\textwidth] (excluded) {Clinically Implausible\\ $\left(n=18\right)$};

                \draw[->, thick] (generate-cases) -- (included);
                \draw[->, thick] (generate-cases) -- ($(generate-cases.south)+(0,-0.44)$) -- node[below]{Excluded}(excluded.east);
            \end{tikzpicture}
            \end{minipage}
        };
        \node[right= of generate-data, draw, rectangle, rounded corners] (evaluate)
        {
            \begin{minipage}{0.45\textwidth}
             \uline{Evaluate \emph{Tak}'s Performance}
             \begin{enumerate}[leftmargin=0cm, itemsep=0cm]
                 \item[] \textbf{Actual: } \emph{Tak} vs. Human Consensus
                 \item[] \textbf{Best: }\emph{Tak} vs. Intended Toxidrome
                 \item[]
                 \item[] \textbf{Benchmarks:}
                 \item[] Human Consensus vs Intended Toxidrome
                 \item[] DT vs Intended Toxidrome 
             \end{enumerate}
            \end{minipage}
        };
        
        \draw[->, thick] (generate-data) -- (evaluate);
    \end{tikzpicture}
    \caption{\textbf{Study flow. Left:} Generation of data using Algorithm \ref{alg:make-patients}.\textbf{Right:} Evaluation of data. Actual denotes actual performance; best, best performance; DT, decision tree, \emph{Tak}, name of combination of probabilistic logic algorithm and knowledge base.}
    \label{fig:study-flow}
\end{figure}

Figure \ref{fig:study-flow} shows the overall organization of our study. $18$ cases were omitted because the human raters could not reach consensus and both said none of them were medically plausible, decreasing the number of cases from $300$ to $282$.

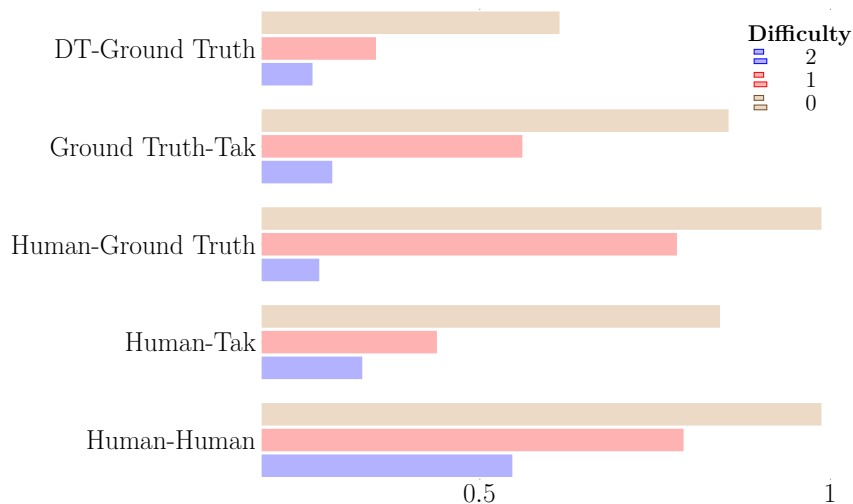
\begin{figure}
     \centering
     \begin{tikzpicture}[scale=0.6]
    \begin{axis}[xbar,xmajorgrids, tick align=inside,
        major grid style={draw=white}, x axis line style = {draw=none},
        axis y line*=left, xtick={0,0.5,1},legend style={draw=none},
        xmajorgrids=true,
        bar width=0.5cm,symbolic y coords={
           Human-Human,Human-Tak,Human-Ground Truth, Ground Truth-Tak, DT-Ground Truth}, y axis line style={opacity=0},tickwidth=0pt,enlarge x limits=true,width=\columnwidth, height=12cm, 
           enlarge y limits=0.1, enlarge x limits=0.1, ytick distance=1,
           legend style={font=\Large},
           tick label style={font=\LARGE}
           ]
            \addlegendimage{empty legend}
            \addlegendentry{\hspace{-1em}\textbf{Difficulty}}
           \foreach \i in {2,1,0} 
           {
                \addplot+[draw=none,discard if not symbolic={difficulty}{\i},] table [
                    col sep=comma,y=type,x=kappa] {combined-consensus-kappas.csv};
                \addlegendentryexpanded{\i}
            }
    \end{axis}
\end{tikzpicture}
     \caption{\textbf{Human and Computer Agreement} Y-axis denotes Cohen's $\kappa$. X-axis denotes pair; Human raters, Intended (actual) labels, and Inferred (Tak) labels. Hue indicates difficulty of presentation. Refers to number of noncanonical symptoms in presentation. \label{fig:summary-of-performace}}
     \end{figure}

Figure \ref{fig:summary-of-performace} summarizes the performances. \emph{Tak}'s peak and actual performance were comparable. As the complexity of the case increased \emph{Tak}'s performance decreased, as did the performance of human experts. In the most difficult cases, \emph{Tak}'s accuracy approached that of the decision tree classifier, approximately half of human performance. In all cases, \emph{Tak's} performance was better than chance ($\kappa =0$) and at least as good as the decision tree, our benchmark for current approaches. 

\subsection{Ideal Performance \label{sec:ideal-performance}}
The inter-rater reliabilities between the toxidromes \emph{Tak} inferred and the intended toxidromes were $\kappa=0.8554$, $0.5614$, and $0.2904$ for difficulty levels $0$,$1$, and $2$, respectively. The inter-rater reliabilities between the consensus of human raters and the intended toxidromes were $\kappa=0.9878$, $0.7818$, and $0.2718$ for difficulty levels $0$,$1$, and $2$, respectively. As the difficulty increased \emph{Tak} developed difficulty distinguishing among the anticholinergic, cholinergic, and sedative-hypnotic toxidromes as well as between the opioid and cholinerigc toxidromes. 

\subsection{Usual Performance}
Our benchmark for usual performance was the inter-rater reliability between the consensus of human raters and the intended toxidromes.  The inter-rater reliability between the consensus of the human raters and the labels predicted by \emph{Tak} was $\kappa=0.8432$, $0.4396$, and $0.3331$, for difficulties $0$,$1$, and $2$, respectively. 

\subsection{Benchmark Performance}
 To compare \emph{Tak}'s performance against other machine learning approaches, we calculated the inter-rater reliability between the ground truth labels and a decision tree classifier. In straightforward presentations (difficulty, $0$) \emph{Tak} outperformed the decision tree ($\kappa_{\textrm{DT}}=0.6144$ vs $\kappa_{Tak}=.8554$). In intermediate difficulty presentations \emph{Tak} outperformed the decision tree ($\kappa_{\textrm{DT}}=0.3527$ vs $\kappa_{Tak}=.5614$). In complex presentations (difficulty=$2$) \emph{Tak} performed comparably to the decision tree ($\kappa_{\textrm{DT}}=0.2622$ vs $\kappa_{Tak}=.2904$). This benchmark was designed to favor the decision trees. 

\subsection{Evaluation of Errors \label{sec:evaluation-of-errors}}
As the difficulty of presentations increased both \emph{Tak} and the human raters decreased in accuracy. This decrease in accuracy reflects the construction of the synthetic data set and the limits of resolution of toxidromes. The synthetic data were constructed to have three levels of difficulty, corresponding to clinical reality. Some patients may ingest or be exposed to a large amount of one substance leading to an unequivocal presentations. Others may ingest or be exposed to a mixture of substances with a variety of stimulating and sedating effects, the balance of which shifts over time as the chemicals are distributed throughout the body and metabolized at different rates. 
   
The authors could find no published formal analysis of the discriminative limits of toxidromes, but it stands to reason that $6$ features may not be able to accurately classify $6$ categories if all features do not have values for all categories.
   
\emph{Tak} confused the anticholinergic and sympathomimetic toxidromes and cholinergic, opioid, and sedative hypnotic toxidromes. This mimics difficulties that medical toxicologists have. The anticholinergic and sympathomimetic toxidromes share overlapping features (increased heart rate, increased blood pressure, and agitated mental status). The cholinergic, opioid, and sedative-hypnotic toxidromes share overlapping features (sedated mental status, and in the case of the cholinergic and opioid toxidromes slowed breathing and small pupils). \emph{Tak} was able to distinguish serotonergic toxicity from the anticholinergic and cholinergic toxidromes because serotonergic toxicity has unique features. Future work can explore the sensitivity of classification to each rule.

\section{Conclusions \label{sec:conclusions}} 
 The goal of this study was to use probabilistic logic to model medical decision-making in medical toxicology. We derived probabilistic logic rules from expert consensus to represent background medical knowledge and constructed a probabilistic logic network, \emph{Tak}. We evaluated \emph{Tak}'s performance on a synthetic data set and compared its performance against the consensus of expert clinicians and a decision tree classifier. 

The misclassification errors made by \emph{Tak} resemble those by humans. The decrease in inter-rater reliability for more difficult cases arose, in part, from the increased chance of the introduction of nondiscriminative physical findings. Toxidromes share overlapping features. For example, an elevated heart rate can be a sign of the anticholinergic or sympathomimetic toxidrome and, indeed, it can be difficult for clinicians to distinguish these two processes without further information.  For example, in a poisoned patient the heart rate and blood pressure usually move in tandem, both rising or both falling. This collinearity is such a hallmark of poisoned patients that its absence can prompt medical toxicologist to consider nontoxicological causes of the patient's condition.

\emph{Tak}'s error rate needs improvement before clinical use. Even if \emph{Tak} never reaches accuracy comparable to human physicians across all levels of difficulty, it could be used to automate the processing of more routine cases, where it's performance is comparable to humans, freeing up physician time to deal with more complex cases.  
 
 The most significant limitation of this paper is the use of synthetic rather than actual data to evaluate our approach. We used synthetic data because no clinical data set currently exists. We took steps to generate realistic data, creating cases at $3$ levels of complexity to capture the heterogeneity of clinical data. We reviewed all cases with medical toixcologists for plausibility. Our data set dotes not fully evaluate \emph{Tak's} performance. It would require $\left(3^5\right) \cdot 4^5$ unique cases to fully evaluate all possible patient presentations, even clinically implausible ones. Our evaluation of actual performance is also limited by the omission of cases on whom the raters could not reach consensus.  The full value of our approach in helping physicians treat poisoned patients will not be known until our approach can be evaluated on actual clinical data. 
 
 For most variables, empiric probability distributions were not available. This renders the absolute values of the calculated posterior probabilities uninformative even if the relative magnitude is still informative. 
 
 Physicians must trust an AI-based system to include it in their evaluation and treatment of patients. An algorithm can earn that trust through proficiency on complex cases and transparency. \emph{Tak} demonstrates transparent clinical reasoning. This transparency, if preserved in more accurate models, may remove barriers to the use of AI approaches in clinical decision making.  Even if a more detailed analysis of the limits of PLNs suggests a unimprovably poor performance on complex cases, a transparent AI-system may be useful by automating aspects routine cases and in doing so freeing up expert time for more complicate cases.  
 
 The main contribution of this paper is the demonstration that probabilistic logic networks can model toxicologic knowledge in a way that transparently mimic physician thought. An additional contribution of this paper is that it is one of the first, to the authors' knowledge, to quantify the inter-rater reliability of physicians in diagnosing a specific type of poisoned patient. Yet another contribution of this paper is the development of a data set that unsupervised or weakly supervised techniques could use to explore other ways of representing knowledge in this domain. 

\addtolength{\textheight}{-.2cm} 
\bibliographystyle{ieeetr}
\bibliography{sample}

\newpage
\section*{Supplemental Material}

 In the Supplemental Material we provide further background on the Mathematics of Probabilisitc Logic Networks, and our Generation of Simulated Patients. 

\subsection*{Mathematics of Probabilistic Logic Networks}
Equation \ref{eq:pnl} shows how the probability associated with a query $Q$ is related to the logical rules and their associated probabilities. The sum ranges over all worlds in which the supplied facts, $F$, and stated rules, $R$, imply that the query $Q$ is true. The products range over all worlds. The first product calculates the joint probability of the supplied facts being true. The second product calculates the joint probability of other facts being false. In our case, $R$ is the set of rules specifying the relationships between signs and toxidromes. The supplied facts, $F$, correspond to signs of a particular patient. The query $Q$ is for each toxidrome. The metaquery or directive is to find the query (toxidrome) that maximizes Equation \ref{eq:pnl}. 
      
     \begin{equation}
         P\left(Q\right) = \sum_{F\cup R\models Q} \prod_{f\in F} p\left(f\right) \prod_{f\not\in F} 1-p\left(f\right) \label{eq:pnl}
     \end{equation}

\subsection*{Generation of Simulated Patients.}
    Algorithm \ref{alg:make-patients} describes how we generated each patient presentation from an intended and distractor toxidrome according to the mixing parameter $k$. 

\begin{algorithm}
\caption{Generation of simulated toxidrome \label{alg:make-patients}}
\begin{algorithmic}
\Require $n \gets 5$\newline \Comment{Max signs per presentation}
\Require $0 \leq k \leq n$ \Comment{difficulty}
\Require  $\{t\}\Leftarrow$ \{sympathomimetic, anticholinergic, cholinergic, sedative_hypnotic, opioid, serotonin_toxicity\} \Comment{toxidromes}
\Require  $\{t,s,v\} \Leftarrow \{\left(t_i,s_{ij},v_{ijk}\right)\}$\newline\Comment{classic values for each sign in a toxidrome}\newline 
\Ensure $\{p\} \Leftarrow \{s_k,v_k\}$\newline \Comment{set of sign, value pairs}\newline
\State  $t_i \Leftarrow \textrm{random.choice}\left(\{t\}\right)$\Comment{intended toxidrome} 
\State $ \hat{t}_i \Leftarrow \textrm{random.choice}\left(\{t\}-t_i\right)\; i \neq j$\newline \Comment{distractor toxidrome}

\State presentation $\gets \{\}$
\While{generated = false}
\State chose $\left(5-k\right) \; \{s,v\}$ pairs from $\{\left(t=t_i,s_{ij},v_{ijk}\right)\}$
\State presentation $\gets$ pairs
\State chose $k\; \{s,v\} $ pairs from $\{\left(t=\hat{t}_i,s_{ij},v_{ijk}\right)\}$
\State presentation $\gets$ pairs 
\EndWhile
\end{algorithmic}
\end{algorithm}

Table \ref{tab:dist-of-cases} shows the distribution of simulated presentations across intended toxidromes and difficulty.

\begin{table}[t]
\centering
\begin{tabular}{@{}llll@{}}
\toprule
 & \multicolumn{3}{l}{Difficulty} \\ \midrule
& \multicolumn{1}{c}{0} & \multicolumn{1}{c}{1} & \multicolumn{1}{c}{2} \\ \cmidrule(l){2-4} 
Anticholinergic & 14 & 14 & 14 \\
Cholinergic & 19 & 26 & 16 \\
Opioid & 25 & 11 & 13 \\
Sedative-Hypnotic & 17 & 13 & 24 \\
Serotonin Toxicity & 10 & 21 & 20 \\
Sympathomimetic & 15 & 15 & 15 \\ \bottomrule
\end{tabular}
\caption{\textbf{Distribution of Simulated Presentations.}\label{tab:dist-of-cases}}
\end{table}

We implemented this algorithm in the Python programming language. We implementated the logical rules in ProbLob, a dialect of Prolog\cite{de2007problog}.

\end{document}